\documentclass{bmcart}

\usepackage[utf8]{inputenc} 
\usepackage{array}
\newcolumntype{L}[1]{>{\raggedright\let\newline\\\arraybackslash\hspace{0pt}}m{#1}}
\newcolumntype{C}[1]{>{\centering\let\newline\\\arraybackslash\hspace{0pt}}m{#1}}
\newcolumntype{R}[1]{>{\raggedleft\let\newline\\\arraybackslash\hspace{0pt}}m{#1}}

\usepackage[export]{adjustbox}
\usepackage{multirow}
\usepackage{url}

\def\includegraphics{}

\startlocaldefs
\endlocaldefs

\begin{document}

\begin{frontmatter}

\begin{fmbox}
\dochead{Research}


\title{Monitoring stance towards vaccination in Twitter messages}


\author[
   addressref={RU},                   
   email={f.kunneman@let.ru.nl}   
]{\inits{FA}\fnm{Florian} \snm{Kunneman}}
\author[
   addressref={RIVM},
   email={mattijs.lambooij@rivm.nl}
]{\inits{M}\fnm{Mattijs} \snm{Lambooij}}
\author[
   addressref={RIVM},
   email={albert.wong@rivm.nl}
]{\inits{A}\fnm{Albert} \snm{Wong}}
\author[
   addressref={RU,MEERTENS},
   email={a.vandenbosch@let.ru.nl}
]{\inits{A}\fnm{Antal} \snm{van den Bosch}}
\author[
   addressref={RIVM},
   email={liesbeth.mollema@rivm.nl}
]{\inits{L}\fnm{Liesbeth} \snm{Mollema}}


\address[id=RU]{
  \orgname{Radboud University}, 
  \street{Erasmusplein 1},                     %
  \postcode{6525 HT}                                
  \city{Nijmegen},                              
  \cny{The Netherlands}                                    
}
\address[id=RIVM]{%
  \orgname{Dutch National Institute for Public Health and Environment},
  \street{Antonie van Leeuwenhoeklaan 9},
  \postcode{3721 MA},
  \city{Bilthoven},
  \cny{The Netherlands}
}
\address[id=MEERTENS]{%
  \orgname{Dutch National Institute for Public Health and Environment},
  \street{Antonie van Leeuwenhoeklaan 9},
  \postcode{3721 MA},
  \city{Bilthoven},
  \cny{The Netherlands}
}



\end{fmbox}


\begin{abstractbox}

\begin{abstract} 
{\bf Background:} We developed a system to automatically classify stance towards vaccination in Twitter messages, with a focus on messages with a negative stance.
  Such a system makes it possible to monitor the ongoing stream of messages on social media, offering actionable insights into public hesitance with respect to vaccination. 
   For Dutch Twitter messages that mention vaccination-related key terms, we annotated their stance and feeling in relation to vaccination (provided that they referred to this topic). Subsequently, we used these coded data to train and test different machine learning set-ups. With the aim to best identify messages with a negative stance towards vaccination, we compared set-ups at an increasing dataset size and decreasing reliability, at an increasing number of categories to distinguish, and with different classification algorithms. 
   
  
\noindent {\bf Results:} We found that Support Vector Machines trained on a combination of strictly and laxly labeled data with a more fine-grained labeling yielded the best result, at an F1-score of $0.36$ and an Area under the ROC curve of $0.66$, outperforming a rule-based sentiment analysis baseline that yielded an F1-score of $0.25$ and an Area under the ROC curve of $0.57$.
  
\noindent {\bf Conclusion:} The outcomes of our study indicate that stance prediction by a computerized system only is a challenging task. Our analysis of the data and behavior of our system suggests that an approach is needed in which the use of a larger training dataset is combined with a setting in which a human-in-the-loop provides the system with feedback on its predictions.  
%
%
\end{abstract}




\end{abstractbox}
%

\end{frontmatter}



\section*{Background}
In the light of increased vaccine hesitance in various countries, consistent monitoring of public beliefs and opinions about the national immunization program is important. Besides performing qualitative research and surveys, real-time monitoring of social media data about vaccination is a valuable tool to this end. The advantage is that one is able to detect and respond to possible vaccine concerns in a timely manner, that it generates continuous data and that it consists of unsolicited, voluntary user-generated content. 

Several studies that analyse tweets have already been conducted, providing  
insight in the content that 
was tweeted most during the 2009 H1N1 outbreak \cite{Chew+10}, the information flow between users with a certain sentiment during this outbreak \cite{Salathe+11}, or trends in tweets that convey, for example, the worries on efficacy of HPV vaccines \cite{Du+17,Massey+16}. While human coders are best at deploying world knowledge and interpreting the intention behind a text, manual coding of tweets is laborious. The above-mentioned studies therefore aimed at developing and evaluating a system to code tweets automatically. There are several systems in place that make use of this automatic coding. The Vaccine Confidence Project \cite{Larson+13} is a real-time worldwide internet monitor for vaccine concerns. The Europe Media Monitor (EMM) \cite{Linge+09} was installed to support EU institutions and Member State organizations with, for example, the analysis real-time news for medical and health-related topics and with early warning alerts per category and country. MEDISYS, derived from the EMM and developed by the Joint Research Center of the European Commission \cite{Rortais+10}, is a media monitoring system providing event-based surveillance to rapidly identify potential public health threats based on information from media reports. 

These systems cannot be used directly for the Netherlands because they do not contain search words in Dutch, are missing an opinion-detection functionality, or do not include categories of the proper specificity. Furthermore, opinions towards vaccination are contextualized by national debates rather than a multinational debate \cite{Becker+16}, which implies that a system for monitoring vaccination stance on Twitter should ideally be trained and applied to tweets with a similar language and nationality. Finally, by creating an automatic system for mining public opinions on vaccination concerns, one can continue training and adapting the system. We therefore believe it will be valuable to build our own system.
Besides analysing the content of tweets, several other applications that use social media with regard to vaccination have been proposed. They, for example, use data about internet search activity and numbers of tweets as a proxy for (changes in) vaccination coverage or for estimating epidemiological patterns. Huang et al. \cite{Huang+17} found a high positive correlation between reported influenza attitude and behavior on Twitter and influenza vaccination coverage in the US. In contrast, Aquino et al. \cite{Aquino+17} found an inverse correlation between Mumps, Measles, Rubella (MMR) vaccination coverage and tweets, Facebook posts and internet search activity about autism and MMR vaccine in Italy. This outcome was possibly due to a decision of the Court of Justice in one of the regions to award vaccine-injury compensation for a case of autism. Wagner, Lampos, Cox and Pebody \cite{Wagner+18} assessed the usefulness of geolocated Twitter posts and Google search as source data to model influenza rates, by measuring their fit to the traditional surveillance outcomes and analyzing the data quality. They find that Google search could be a useful alternative to the regular means of surveillance, while Twitter posts are not correlating well due to a lower volume and bias in demographics. Lampos, de Bie and Christianinni \cite{Lampos+10} also make use of geolocated Twitter posts to track academics, and present a monitoring tool with a daily flu-score based on weighted keywords.  

Various studies \cite{Nagar+14, Kim+13, Signorini+11} show that estimates of influenza-like illness symptoms mentioned on Twitter can be exploited to track reported disease levels relatively accurately. However, other studies \cite{Vasterman+13, Mollema+15} showed that this was only the case when looking at severe cases (e.g. hospitalizations, deaths) or only for the start of the epidemic when interest from journalists was still high. 

Other research focuses on detecting discussion communities on vaccination in Twitter \cite{Bello-orgaz+17} or analysing semantic networks \cite{Kang+17} to identify the most relevant and influential users as well as to better understand complex drivers of vaccine hesitancy for public health communication. Tangherlini et al. \cite{Tangherlini+16} explore what can be learned about the vaccination discussion from the realm of “mommy blogs”: parents posting messages about children’s health care on forum websites. They aim to obtain insights in the underlying narrative frameworks, and analyse the topics of the messages using Latent Dirichlet Allocation (LDA) \cite{Blei+03}. They find that the most prominent frame is a focus on the exemption of one’s child from receiving a vaccination in school. The motivation against vaccination is most prominently based on personal belief about health, but could also be grounded in religion. Surian et al. \cite{Surian+16} also apply topic modeling to distinguish dominant opinions in the discussion about vaccination, and focus on HPV vaccination as discussed on Twitter. They find a common distinction between tweets reporting on personal experience and tweets that they characterize as `evidence’ (statements of having had a vaccination) and `advocacy’ (statements that support vaccination). 

Most similar to our work is the study by Du, Xu, Song, Liu and Tao \cite{Du+17}. With the ultimate aim to improve the vaccine uptake, they applied supervised machine learning to analyse the stance towards vaccination as conveyed on social media. Messages were labeled as either related to vaccination or unrelated, and, when related, as ‘positive’, ‘negative’ or ‘neutral’. The ‘negative’ category was further broken down into several considerations, such as ‘safety’ and ‘cost’. After having annotated 6,000 tweets, they trained a classifier on different combinations of features, obtaining the highest macro F1-score (the average of the separate F1-scores for each prediction category) of $0.50$ and micro F1-score (the F1-score over all predictions) of $0.73$. Tweets with a negative stance that point to safety risks could best be predicted, at an optimal F1 score of $0.75$, while the other five sub-categories with a negative stance were predicted at an F1 score below $0.5$ or even $0.0$. 

Like Du et al. \cite{Du+17}, we focus on analysing sentiment about vaccination using Twitter as a data source and applying supervised machine learning approaches to extract public opinion from tweets automatically. In contrast, in our evaluation we focus on detecting messages with a negative stance in particular. Accurately monitoring such messages helps to recognize discord in an early stage and take appropriate action. We do train machine learning classifiers on modeling other categories than the negative stance, evaluating whether this is beneficial to detecting tweets with a negative stance. For example, we study whether it is beneficial to this task to model tweets with a positive and neutral stance as well. We also inquire whether a more fine-grained categorization of sentiment (e.g.: worry, relief, frustration and informing) offers an advantage. Apart from comparing performance in the context of different categorizations, we compare different machine learning algorithms and compare data with different levels of annotation reliability. Finally, the performance of the resulting systems is compared to regular sentiment analysis common to social media monitoring dashboards. At the public health institute in the Netherlands, we make use of social media monitoring tools offered by Coosto\footnote{\url{https://www.coosto.com/en}}. For defining whether a message is positive, negative or neutral with regard to vaccination, this system makes use of the presence or absence of positive or negative words in the messages. We believe that we could increase the sensitivity and specificity of the sentiment analysis by using supervised machine learning approaches trained on a manually coded dataset. The performance of our machine learning approaches is therefore compared to the sentiment analysis that is currently applied in the Coosto tool.

\section*{Implementation}

We set out to curate a corpus of tweets annotated for their stance towards vaccination, and to employ this corpus to train a machine learning classifier to distinguish tweets with a negative stance towards vaccination from other tweets. In the following, we will describe the stages of data acquisition, from collection to labeling. 

\subsection*{Data collection}
We queried Twitter messages that refer to a vaccination-related key term from TwiNL \footnote{\url{https://twinl.surfsara.nl/}}, a database with IDs of Dutch Twitter messages from January 2012 onwards \cite{TjongKimSang+13}. In contrast to the open Twitter Search API\footnote{\url{https://developer.twitter.com/en/docs/tweets/search/api-reference}}, which only allows one to query tweets posted within the last seven days, TwiNL makes it possible to collect a much larger sample of Twitter posts, ranging several years. 

We queried TwiNL for different key terms that relate to the topic of vaccination in a five-year period, ranging from January 1, 2012 until February 8, 2017. Query terms that we used were the word `vaccinatie’ (Dutch for `vaccination’) and six other terms closely related to vaccination, with and without a hashtag (`\#’). Among the six words is  `rijksvaccinatieprogramma’, which refers to the vaccination programme in The Netherlands. An overview of all query terms along with the number of tweets that could be collected based on them is displayed in Table \ref{tab:queried}.  

We collected a total of 96,566 tweets from TwiNL, which we filtered in a number of ways. First, retweets were removed, as we wanted to focus on unique messages.\footnote{Although original content of the sender could be added to retweets, this was only manifested in a small part of the retweets in our dataset. It was therefore most effective to remove them.} This led to a removal of 31\% of the messages. Second, we filtered out messages that contain a URL. Such messages often share a news headline and include a URL to refer to the complete news message. As a news headline does not reflect the stance of the person who posted the tweet, we decided to apply this filtering step. It is likely that part of the messages with a URL do include a message composed by the sender itself, but this step helps to clean many unwanted messages. Third, we removed messages that include a word related to animals and traveling (`dier’, animal; `landbouw’, agriculture; and `teek’, tick), as we strictly focus on messages that refer to vaccination that is part of the governmental vaccination program. 27,534 messages were left after filtering. This is the data set that is used for experimentation.  
%

\small
\begin{table}[t]
\caption{Overview of the number of Twitter messages that were queried from TwiNL and filtered, from the period between January 2012 and February 2017 . `URLs' refers to tweets with a URL. `blacklist' refers to words related to animal vaccination and vaccination related to travelling: `dier' (animal), `landbouw' (agriculture),  and `teek' (tick).}
      \begin{tabular}{llrrrr}
        \hline
         Query term & Query term & Before & Excluding & Excluding & Excluding  \\ 
         (original) & (English) & filtering & retweets & URLs & blacklist \\ \hline
		 Vaccinatie & Vaccination & 30,730 & 20,677 & 8,872 & 8,818 \\
         Vaccin & Vaccine & 21,614 & 16,046 & 4,154 & 4,121 \\
         Vaccineren & Vaccinate & 20,689 & 11,904 & 4,682 & 4,653 \\
		 Rijksvaccinatieprogramma & Gov. vacc. programme & 1,151 & 520 & 160 & 158 \\
         Vaccinatieprogramma & Vacc. programme & 644 & 407 & 121 & 120 \\
		 Inenting & Inoculation & 8,597 & 7,093 & 4,046 & 4,038 \\
		 Inenten & Inoculate & 13,141 & 9,535 & 5,640 & 5,626 \\ \hline 
		 \multicolumn{2}{c}{Total} & 96,566 & 66,182 & 27,675 & 27,534 \\ \hline
      \end{tabular}
      \label{tab:queried}
\end{table}

\subsection*{Data annotation}

\begin{table}[h!]
\caption{Specification of the annotation categories}
\begin{scriptsize}
\begin{tabular}{ L{1.3cm} | L{1.5cm} | L{4cm} | L{4cm} }
\hline
Category type &	Category & Definition & Example tweet (translated from Dutch) \\ \hline
\multirow{3}{*}{Relevance} &	Relevant & If the message is about (information about) human vaccination or expresses an opinion about human vaccination. & “By the way I do not accuse people who are against vaccination. It is just that they should not imply that the measles are so harmless.” \\ \cline{2-4}
 &	Relevant abroad & If the message is relevant and is about an event related to vaccination or an outbreak of vaccine preventable disease that happens abroad. & “Have you seen the Danish detective on chronic fatigue after HPV-vaccination?” \\ \cline{2-4}
 &	Irrelevant & If the message is not about human vaccination. & “Lethal virus has been fatal to at least twelve rabbits in Hellevoetsluis. Veterinarians sound the alarm: get inoculation \#ADRD \#VoornePutten” \\ \hline
\multirow{3}{*}{Subject} &	Vaccine & If the message contains an expression about the vaccine.  & “Rutte: pastors please encourage inoculation measles” \\ \cline{2-4}
 &	Disease & If the message contains an expression about the disease. & “I am not happy. I have the chickenpox, which is not in the governmental vaccination program.” \\ \cline{2-4}
 &	Vaccine and disease & If the message contains an expression about both the vaccine and disease. & “I think the whooping-cough disease is rather significant, too bad the vaccine does not have much effect.” \\ \hline
\multirow{4}{*}{Stance} &	Positive & If one is positive with regard to vaccination and/or believes the vaccine preventable disease is severe. & “To inoculate against the measles is at least better than not inoculating. The reformed church is also divided about this.” \\ \cline{2-4}
 &	Negative & If one is negative towards vaccination and/or believes the vaccine preventable disease is not severe. & “Did you ever check the number of casualties as a result of vaccination? Now those are really in vain. One does not die from \#measles” \\ \cline{2-4}
 &	Neutral & If one takes a neutral stance towards vaccination and if one only wants to inform others. & “[anonymized name] : inoculating at home \#measles at \#refo’s” \\ \cline{2-4}
 &	Not clear & If from the message it is not clear whether one is positive or negative, if both polarities are present, or if the message is about a related topic such as information about vaccination. & “Facts and opinions related to \#HPV vaccination: why is it almost impossible to find them on the website of \#RIVM?” \\ \hline
\multirow{5}{*}{Sentiment} &	Informative & If one wants to inform others.  & “GGGD\_Utrecht: Today the GG\&GD will start vaccinating all 9-year olds against DTP and BMR. This applies to 3395 kids in Utrecht!” \\ \cline{2-4}
 &	Anger, frustration & If one is angry about people who vaccinate or do not vaccinate. & “Measles epidemic in the \#biblebelt. Incomprehensible that the love for God can be greater than the love for one’s own child.” \\ \cline{2-4}
 &	Worry, fear, doubts & If one is worried about side-effects of the vaccine or about the severity of the disease; if one has doubts to vaccinate. & “I will watch zorg.nu in a bit. This time I am doubtful once more as to whether I should have my youngest daughter inoculated against cervical cancer.” \\ \cline{2-4}
 &	Relieved & If one is relieved that the vaccination has been administered or that he/she recovered from the disease. & “I am happy that the vaccination is over with.” \\ \cline{2-4}
 &	Other  & If one expresses another sentiment than those mentioned above, such as humor, sarcasm (see example), personal experience, question raised, or minimized risks. & “What a genius idea of the doctor to vaccinate me for yellow fever, polio, meningitis, and hepatitis A, all in once! Bye bye weekend.. ” \\ \hline
\end{tabular}
\end{scriptsize}
\label{tab:examples}
\end{table}

The stance towards vaccination was categorized into `Negative’, `Neutral’, `Positive’ and `Not clear’. The latter category was essential, as some posts do not convey enough information about the stance of the writer. In addition to the four-valued {\bf stance} classes we included separate classes grouped under {\bf relevance}, {\bf subject} and {\bf sentiment} as annotation categories. With these additional categorizations we aimed to obtain a precise grasp of all possibly relevant tweet characteristics in relation to vaccination, which could help in a machine learning setting.\footnote{We give a full overview of the annotated categories, to be exact about the decisions made by the annotators. However, we did not include all annotation categories in our classification experiment. A motivation will be given in the `Data categorization' section.} 

The {\bf relevance} categories were divided into `Relevant’, `Relevant abroad’ and `Irrelevant’. Despite our selection of vaccination-related keywords, tweets that mention these words might not refer to vaccination at all. A word like `vaccine’ might be used in a metaphorical sense, or the tweet might refer to vaccination of animals. 

The {\bf subject} categorization was included to describe what the tweet is about primarily: `Vaccine’, `Disease’ or `Both’. We expected that a significant part of the tweets would focus on the severeness of a disease when discussing vaccination. Distinguishing these tweets could help the detection of the stance as well. 

Finally, the {\bf sentiment} of tweets was categorized into `Informative’, `Angry/Frustration’, `Worried/Fear/Doubts’, `Relieved’ and `Other’, where the latter category lumps together occasional cases of humor, sarcasm, personal experience, and question raised. These categories were based on the article by \cite{Chew+10}, and emerged from analysing their H1N1-related tweets. The `Informative’ category refers to a typical type of message in which information is shared, potentially in support of a negative or positive stance towards vaccination. If the message contained more than one sentiment, the first sentiment identified was chosen. Table \ref{tab:examples} shows examples of tweets for the above-mentioned categories.

We aimed at a sufficient number of annotated tweets to feed a machine learning classifier with. The majority of tweets were annotated twice. We built an annotation interface catered to the task. Upon being presented with the text of a Twitter post, the annotator was first asked whether the tweet was relevant. In case it was deemed relevant, the tweet could be annotated for the other categorizations. Otherwise, the user could click `OK’ after which he or she was directly presented with a new Twitter post. The annotator was presented with sampled messages that were either not annotated yet or annotated once. We ensured a fairly equal distribution of these two types, so that most tweets would be annotated twice.  	

As annotators, we hired four student assistants and additionally made use of the Radboud Research Participation System.\footnote{\url{https://radboud.sona-systems.com}} 
We asked participants to annotate for the duration of an hour, in exchange for a voucher valued ten Euros, or one course credit. Before starting the annotation, the participants were asked to read the annotation manual, with examples and an extensive description of the categories, and were presented with a short training round in which feedback on their annotations was given. The annotation period lasted for six weeks. We stopped when the number of applicants dropped.  

A total of 8,259 tweets were annotated, of which 6,472 were annotated twice (78\%).\footnote{The raw annotations by tweet identifier can be downloaded from \url{http://cls.ru.nl/~fkunneman/data_stance_vaccination.zip}} 65 annotators joined in the study, with an average of $229.5$ annotated tweets per person. The number of annotations per person varied considerably, with $2,388$ tweets coded by the most active annotator. This variation is due to the different ways in which annotators were recruited: student-assistants were recruited for several days, while participants recruited through the Radboud Research Participation System could only join for the duration of an hour. 

\begin{table}[t]
\caption{Agreement scores for all four categorizations; mutual F-score is reported by category.}
\begin{scriptsize}
\begin{tabular}{ l | l l l l l l l l }
    \hline
	& \multicolumn{2}{c}{Relevance} & \multicolumn{2}{c}{Subject} & \multicolumn{2}{c}{Stance} & \multicolumn{2}{c}{Sentiment} \\ \hline
	Percent agreement & \multicolumn{2}{r}{0.71} & \multicolumn{2}{r}{0.70} & \multicolumn{2}{r}{0.54} & \multicolumn{2}{r}{0.54} \\ 
	Krippendorff's Alpha & \multicolumn{2}{r}{0.27} & \multicolumn{2}{r}{0.29} & \multicolumn{2}{r}{0.35} & \multicolumn{2}{r}{0.34} \\ 
	Mutual F-score & Relevant & 0.81 & Vaccine & 0.79 & Negative & 0.42 & Worry, fear, doubts & 0.21 \\ 
	& Relevant abroad & 0.40 & Disease & 0.06 & Neutral & 0.23 & Anger, frustration & 0.50 \\
	& Irrelevant & 0.42 & Vaccine and disease & 0.49 & Positive & 0.64 & Informative & 0.49 \\
	& & & & & Not clear & 0.31 & Relieved & 0.19 \\
	& & & & & & & Other & 0.20 \\
	\hline
\end{tabular}
\end{scriptsize}
\label{tab:agreement}
\end{table}

We calculated inter-annotator agreement by Krippendorff's Alpha \cite{Hayes+07}, which accounts for different annotator pairs and empty values. To also zoom in on the particular agreement by category, we calculated mutual F-scores for each of the categories. This metric is typically used to evaluate system performance by category on gold standard data, but could also be applied to annotation pairs by alternating the roles of the two annotators between classifier and ground truth. A summary of the agreement by categorization is given in Table \ref{tab:agreement}. While both the Relevance and Subject categorizations are annotated at a percent agreement of $0.71$ and $0.70$, their agreement scores are only fair, at $\alpha=0.27$ and $\alpha=0.29$. The percent agreement on Stance and Sentiment, which carry more categories than the former two, is $0.54$ for both. Their agreement scores are also fair, at $\alpha=0.35$ and $\alpha=0.34$. The mutual F-scores show marked differences in agreement by category, where the categories that were annotated most often typically yield a higher score. This holds for the Relevant category ($0.81$), the Vaccine category ($0.79$) and the Positive category ($0.64$). The Negative category yields a mutual F-score of $0.42$, which is higher than the more frequently annotated categories Neutral ($0.23$) and Not clear ($0.31$). We found that these categories are often confused. After combining the annotations of the two, the stance agreement would be increased to $\alpha=0.43$. 

The rather low agreement over the annotation categories indicates the difficulty of interpreting stance and sentiment in tweets that discuss the topic of vaccination. We therefore proceed with caution to categorize the data for training and testing our models. The agreed upon tweets will form the basis of our experimental data, as was proposed by Jakubiçek, Kovar and Rychly \cite{Jakubicek+14}, while the other data is added as additional training material to see if the added quantity is beneficial to performance. We will also annotate a sample of the agreed upon tweets, to make sure that these data are reliable in spite of the low agreement rate. 

\subsection*{Data categorization}
The labeled data that we composed based on the annotated tweets are displayed in Table \ref{tab:labeled_data}.  We combined the Relevant and Relevant abroad categories into one category (`Relevant’), as only a small part of the tweets was annotated as Relevant abroad. We did not make use of the {\bf subject} annotations, as a small minority of the tweets that were relevant referred a disease only. For the most important categorization, {\bf stance}, we included all annotated labels. Finally, we combined part of the more frequent sentiment categories with Positive.

We distinguish three types of labeled tweets: `strict’, `lax’ and `one’. The strictly labeled tweets were labeled by both annotators with the same label. The lax labels describe tweets that were only annotated with a certain category by one of the coders. The categories were ordered by importance to decide on the lax labels. For instance, in case of the third categorization, Negative was preferred over Positive, followed by Neutral, Not clear and Irrelevant. If one of the annotators labeled a tweet as Positive and the other as Neutral, the lax label for this tweet is Positive. In table \ref{tab:labeled_data}, the categories are ordered by preference as imposed on the lax labeling. The `one' labeling applies to all tweets that were annotated by only one annotator. Note that the total counts can differ between label categorizations due to the lax labeling: the counts for Positive labels in the Polarity + sentiment labeling (Positive + Frustration, Positive + Information and Positive + other) do not add up to the count of the Positive label in the Polarity labeling.   

With the `strict’, `lax’ and `one’ labeling, we end up with four variants of data to experiment with: only strict, strict + lax, strict + one and strict + lax + one. The strict data, which are most reliable, are used in all variants. By comparing different combinations of training data, we test whether the addition of less reliably labeled data (lax and/or one) boosts performance. 

\begin{table}[t!]
\caption{Overview of data set (the cells indicate the number of examples per label and data type)}
\begin{tabular}{ l|l|l|l|l|l }
\hline
	 &  & \multicolumn{4}{c}{Training data} \\
	 & & Strict & Strict + & Strict + & Strict + \\ 
    Labeling & Labels & & Lax & One & Lax + \\
     & & & & & One \\ \hline
	Binary & Negative & 343 & 1,188 & 534 & 1,379 \\ 
	 & Other & 2,543 & 5,358 & 4,074 & 6,889 \\ \hline
	Irrelevance filter & Negative & 343 & 1,188 & 534 & 1,379 \\ 
	 & Irrelevant & 633 & 633 & 1,077 & 1,077 \\ 
	 & Other & 1,910 & 4,725 & 2,997 & 5,812 \\ \hline
	Polarity & Negative & 343 & 1,188 & 534 & 1,379 \\ 
	 & Positive & 1,312 & 2,693 & 1,835 & 3,216 \\ 
	 & Neutral & 345 & 1,271 & 623 & 1,549 \\ 
	 & Not Clear & 253 & 761 & 539 & 1,047 \\ 
	 & Irrelevant & 633 & 633 & 1,077 & 1,077 \\ \hline
	Polarity + Sentiment & Negative & 343 & 1,188 & 534 & 1,379 \\ 
	 & Positive + Frustration & 392 & 726 & 560 & 894 \\ 
	 & Positive + Information & 300 & 1,084 & 513 & 1,297 \\ 
	 & Positive + Other & 620 & 879 & 762 & 1,021 \\ 
	 & Neutral & 345 & 1,271 & 623 & 1,549 \\ 
	 & Not Clear & 253 & 761 & 539 & 1,047 \\ 
	 & Irrelevant & 633 & 633 & 1,077 & 1,077 \\ \hline
\end{tabular}
\label{tab:labeled_data}
\end{table}

The four labelings have an increasing granularity, where the numbers of examples for the Negative category are stable across each labeling. In the first labeling, these examples are contrasted with any other tweet. It hence comprises a binary classification task. In the second labeling, irrelevant tweets are indicated in a separate category. The Other class here represents all relevant tweets that do not convey a negative stance towards vaccination. In the third labeling, this class is specified as the {\bf stance} categories Positive, Neutral and Not clear. In the fourth labeling, the Positive category, which is the most frequent polarity class, is further split into `Positive + frustration’, `Positive + Information’ and `Positive + Other’. Positivity about vaccination combined with a frustration sentiment reflects tweets that convey frustration about the arguments of people who are negative about vaccination (e.g.: "I just read that a 17 year old girl died of the measles. Because she did not want an inoculation due to strict religious beliefs. -.- \#ridiculous"). The Positive + Information category reflects tweets that provide information in favor of vaccination, or combined with a positive stance towards vaccination (e.g.: "\#shingles is especially common with the elderly and chronically diseased. \#vaccination can prevent much suffering. \#prevention").\footnote{The tweet IDs and their labels can be downloaded from \url{http://cls.ru.nl/~fkunneman/data_stance_vaccination.zip}} 

In line with Kovár, Rychlý and Jakubíček \cite{Kovar+14}, we evaluate system performance only on the reliable part of the annotations - the instances labeled with the same label by two annotators. As the overall agreement is not sufficient, with Krippendorff's Alpha ranging between $0.27$ and $0.35$, the first author annotated 300 tweets sampled from the strict data (without knowledge of the annotations) to rule out the possibility that these agreed upon annotations are due to chance agreement. Comparing these new annotations to the original ones, the Negative category and the Positive category are agreed upon at mutual F-scores of $0.70$ and $0.81$. The percent agreement on the binary classification scheme (e.g.: Negative versus Other) is $0.92$, with $\alpha=0.67$, which decreases to $\alpha=0.55$ for the Relevance categorization, $\alpha=0.54$ for the Polarity categorization and $\alpha=0.43$ for the Polarity + Sentiment categorization. We find that instances of a negative and positive stance can be clearly identified by humans, while the labels Neutral and Not Clear are less clear cut. Since it is our focus to model tweets with a negative stance, the agreement on the binary decision between Negative and Other is just sufficient to use for experimentation based on Krippendorff’s \cite{Krippendorff04} remark that “$\alpha \geq .667$ is the lowest conceivable limit” (p.241). In our experimental set-up we will therefore only evaluate our system performance on distinguishing the Negative category from any other category in the strict data.

\subsection*{Experimental Set-up}
For each combination of labeling (four types of labeling) and training data (four combinations of training data) we train a machine learning classifier to best distinguish the given labels. Two different classifiers are compared: Multinomial Naive Bayes and Support Vector Machines (SVM). In total, this makes for 32 variants (4 labelings $\times$ 4 combinations of training data $\times$ 2 classifiers). All settings are tested through ten-fold cross-validation on the strict data and are compared against two rule-based sentiment analysis baselines and two random baselines. All components of the experimental set-up are described in more detail below.

\subsubsection*{Preprocessing}
To properly distinguish word tokens and punctuation we tokenized the tweets by means of Ucto, a rule-based tokenizer with good performance on the Dutch language, and with a configuration specific for Twitter.\footnote{\url{https://languagemachines.github.io/ucto/}} Tokens were lowercased in order to focus on the content. Punctuation was maintained, as well as emoji and emoticons. Such markers could be predictive in the context of a discussion such as vaccination. To account for sequences of words and characters that might carry useful information, we extracted word unigrams, bigrams, and trigrams as features. Features were coded binary, i.e. set to 1 if a feature is seen in a message and set to 0 otherwise. During training, all features apart from the top 15,000 most frequent ones were removed.   
%
%

\subsubsection*{Machine Learning}
We applied two machine learning algorithms with a different perspective on the data: Multinomial Naive Bayes and SVM. The former algorithm is often used on textual data. It models the Bayesian probability of features to belong to a class and makes predictions based on a linear calculation. Features are naively seen as independent of one another \cite{Hand+01}. In their simplest form, SVMs are binary linear classifiers that make use of kernels. They search for the optimal hyperplane in the feature space that maximizes the geometric margin between any two classes. The advantage of SVMs is that they provide a solution to a global optimization problem, thereby reducing the generalization error of the classifier \cite{Hearst+98}.  

We applied both algorithms by means of the scikit-learn toolkit, a python library that offers implementations of many machine learning algorithms \cite{Pedregosa+11}. To cope with imbalance in the number of instances per label, for Multinomial Naive Bayes we set the Alpha parameter to $0.0$ and muted the fit prior. For SVM, we used a linear kernel with the $C$ parameter set to $1.0$ and a balanced class weight.  
%
%

\subsubsection*{Baselines}
As baselines, we applied two rule-based sentiment analysis systems for Dutch as well as two random baselines. The first rule-based sentiment analysis system is Pattern, an off-the-shelf sentiment analysis system that makes use of a list of adjectives with a positive or negative weight, based on human annotations \cite{Smedt+12}. Sentences are assigned a score between $-1.0$ and $1.0$ by multiplying the scores of their adjectives. Bigrams like `horribly good’ are seen as one adjective, where the adjective `horribly’ increases the positivity score of `good’. We translated the polarity score into the discrete labels `Negative’, `Positive’ and `Neutral’ by using the training data to infer which threshold leads to the best performance on the `Negative’ category.

The second baseline is the sentiment analysis offered by the social media monitoring dashboard Coosto. As Coosto is a commercial product, there is no public documentation on their sentiment analysis tool. 

In addition to these two baselines, we applied two random baselines: predicting the negative class randomly for 50\% of the messages and predicting the negative class randomly for 15\% of the messages. The latter proportion relates to the proportion of vaccination-hesitant tweets in the strictly labeled data on which we test the systems. 

\subsection*{Evaluation}

We evaluate performance by means of ten-fold cross-validation on the strictly labeled data. In each of the folds, 90\% of the strictly labeled data is used as training data, which are complemented with the laxly labeled data and/or the data labeled by one annotator, in three of the four training data variants. Performance is always tested on the strict data. As evaluation metrics we calculate the F1-score and the Area Under the ROC Curve (AUC) on predicting the negative stance towards vaccination in the test tweets.

\section*{Results}

We trained machine learning (ML) classifiers to distinguish Twitter messages with a negative stance towards vaccination, alternating three aspects of the system: the labels to train on, the composition of the training data and the ML algorithm. The results are presented in Table \ref{tab:results}, as the F1-score and AUC of any setting on correctly predicting tweets with a negative stance. Systems with specific combinations of the ML classifier and size of the training data are given in the rows of the table. The four types of labelings are listed in the columns. 

The results show a tendency for each of the three manipulations. Regarding the ML algorithm, SVM consistently outperforms Naive Bayes for this task. Furthermore, adding additional training data, albeit less reliable, generally improves performance. Training a model on all available data (strict + lax + one) leads to an improvement over using only the strict data, while adding only the laxly labeled data is generally better than using all data. Adding only the data labeled by one annotator often leads to a worse performance. With respect to the labeling, the Polarity-sentiment labeling generally leads to the best outcomes, although the overall best outcome is yielded by training an SVM on Polarity labeling with strict data appended by lax data, at an area under the curve score of $0.66$\footnote{We choose to value the AUC over the F1-score, as the former is more robust in case of imbalanced test sets}. 

\begin{table}[t!]
\caption{Machine Learning performance of correctly predicting the label of tweets with a negative stance (Clf = Classifier, NB = Naive Bayes, SVM = Support Vector Machines, AUC = Area under the curve).}
\begin{tabular}{ l l | r r | r r | r r | r r }
\hline
	& & \multicolumn{2}{c|}{Binary} & \multicolumn{2}{c|}{Irrelevance} & \multicolumn{2}{c|}{Polarity} & \multicolumn{2}{c}{Polarity -} \\
    & & & & \multicolumn{2}{c|}{filter} & & & \multicolumn{2}{c}{Sentiment} \\
	Training data & Clf & F1 & AUC & F1 & AUC & F1 & AUC & F1 & AUC \\ \hline
	Strict & NB & 0.14 & 0.53 & 0.15 & 0.54 & 0.24 & 0.56 & 0.30 & 0.60 \\ 
	& SVM & 0.30 & 0.59 & 0.32 & 0.61 & 0.34 & 0.62 & 0.35 & 0.63 \\ 
	Strict + Lax & NB & 0.26 & 0.58 & 0.27 & 0.59 & 0.33 & 0.63 & 0.36 & 0.64 \\ 
	& SVM & 0.33 & 0.63 & 0.34 & 0.63 & 0.36 & \textbf{0.66} & 0.36 & 0.64 \\ 
	Strict + One & NB & 0.13 & 0.53 & 0.15 & 0.54 & 0.24 & 0.57 & 0.27 & 0.59 \\ 
	& SVM & 0.29 & 0.59 & 0.29 & 0.59 & 0.34 & 0.62 & \textbf{0.37} & 0.64 \\ 
	Strict + Lax + One & NB & 0.27 & 0.58 & 0.27 & 0.59 & 0.33 & 0.62 & 0.32 & 0.61 \\ 
	 & SVM & 0.34 & 0.63 & 0.32 & 0.62 & 0.35 & 0.64 & 0.36 & 0.64 \\ \hline
\end{tabular}
\label{tab:results}
\end{table}

\begin{table}[t!]
\caption{Baseline performance of correctly predicting the label of tweets with a negative stance (for comparison, the best ML system is included; Pr = Precision, Re = Recall, AUC = Area under the Curve).}
\begin{tabular}{l | rrrr}
\hline
	 & Pr & Re & F1 & AUC \\ \hline
	Random (50\%) & 0.11 & 0.46 & 0.18 & 0.48 \\ 
	Random (15\%) & 0.12 & 0.15 & 0.13 & 0.50 \\ 
	Pattern & 0.14 & 0.34 & 0.20 & 0.53 \\
	Coosto & 0.20 & 0.31 & 0.25 & 0.57 \\
	Best ML system & 0.29 & 0.47 & 0.36 & 0.66 \\ \hline
\end{tabular}
\label{tab:results_baseline}
\end{table}

\begin{table}[t!]
\caption{Confusion table of the classification of tweets in the best ML setting (SVM trained on Polarity labeling with strict data appended by lax data). The vertical axes give gold standard labels, the horizontal axes give the classifier decisions. Numbers given in bold are accurate classifications.}
\begin{tabular}{ l | l l l l l }
\hline
	& Irrelevant & Negative & Neutral & Positive & Not clear \  \\ \hline
	Irrelevant & \textbf{172} & 17 & 20 & 60 & 25 \\ 
	Negative & 74 & \textbf{161} & 42 & 230 & 57 \\ 
	Neutral & 108 & 37 & \textbf{118} & 133 & 55 \\ 
	Positive & 195 & 103 & 140 & \textbf{832} & 84 \\ 
	Not clear & 84 & 25 & 25 & 57 & \textbf{32} \\ \hline
\end{tabular}
\label{tab:conf_table}
\end{table}

\begin{figure}[t!]
\includegraphics{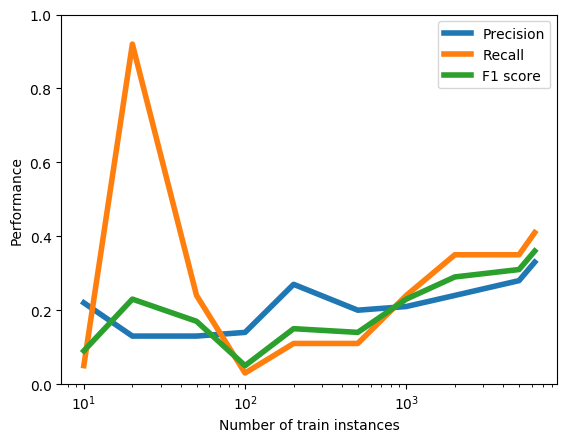}
\caption{\csentence{}Learning curve of the best ML system.}
\label{fig:learning_curve}
\end{figure}

The best reported performance is an F1-score of $0.36$ and an AUC of $0.66$. In comparison to the baselines (Table \ref{tab:results_baseline}), these scores are considerably higher. Nevertheless, there is room for improvement. The performance of the random baselines, with  F1-scores of $0.18$ (50\%) and $0.13$ (15\%), indicates that the minimal performance on this task is rather low. The rule-based sentiment analyses yield better performances, at an F1-score of $0.20$ for Pattern and $0.25$ for Coosto. To analyse the behavior of the best ML system, we present a confusion table of its classifications in Table \ref{tab:conf_table}. The Irrelevant category is most often classified with one of the other categories, while the Positive and Negative categories are the biggest confusables. The classifier is possibly identifying features that denote a stance, but struggles to distinguish positive from negative. 

To gain insight into the potential of increasing the amount of training data, we applied the best ML system (SVM trained on strict and lax data on the polarity labels) on 10\% of the strictly labeled data, starting with a small sample of the data and increasing it to all available data (excluding the test data). The learning curve is presented in Figure \ref{fig:learning_curve}. It shows an improved performance until the last training data is added, indicating that more training data would likely yield better performance.


\subsection*{Comparison machine learning and rule-based sentiment analysis}

A confusion table of the predictions of the best of the two rule-based baselines, Pattern, and the best ML system is displayed in Table \ref{tab:conf_table1}. Only 192 tweets are labeled by both systems as Negative, while the best ML system accounts for almost double this amount and Pattern for three times as much. Comparing the predictions to the gold standard labeling, 99 of the tweets predicted only by the best ML system as Negative are correct (27\%), opposed to 51 that are exclusive to Pattern (8\%). Of the tweets that were classified by both as negative, 63 are correct (33\%). This shows that the approaches have a rather complementary view on tweets with a negative stance. 

\begin{table}[t!]
\caption{Confusion table of the labeled Twitter messages predicted as `Negative’ or another category between Pattern and the best ML system.}
      \begin{tabular}{ll|rr}
        \hline
         & & \multicolumn{2}{c}{Best ML system} \\
         & & Other & Negative \\ \hline
		 \multirow{2}{*}{Pattern} & Other & 1,718 & 372 \\ 
         & Negative & 604 & 192 \\ \hline
      \end{tabular}
      \label{tab:conf_table1}
\end{table}

\begin{table}[t!]
\caption{Confusion table of the unlabeled Twitter messages predicted as ‘Negative’ or another category between Pattern and the best ML system.}
      \begin{tabular}{ll|rr}
        \hline
         & & \multicolumn{2}{c}{Best ML system} \\
         & & Other & Negative \\ \hline
		 \multirow{2}{*}{Pattern} & Other & 8,954 & 2,225 \\ 
         & Negative & 3,383 & 1,015 \\ \hline
      \end{tabular}
      \label{tab:conf_table2}
\end{table}

To gain more insight into the behavior of both approaches, we applied them to 15,577 unlabeled tweets. Table \ref{tab:conf_table2} presents a confusion table with the numbers of tweets that were classified as Negative or another category by both approaches. Again, pattern accounts for the majority of negatively labeled messages, and the overlap is small. Two of the authors validated for a sample of 600 messages whether they actually manifested a negative attitude towards vaccination: 200 messages that were uniquely classified by the best ML system as Negative, 200 messages that were solely labeled as Negative by Pattern and 200 messages that were classified by both systems as Negative. This validation showed the same tendency as for the labeled data, with a higher precision of the best ML system in comparison to Pattern (33.5\% versus 21\% of the messages correctly predicted) and the highest precision when both systems predicted the negative class (36\%).  

The complementary view on tweets with a negative stance between the best ML system and rule-based sentiment analysis becomes clear from their differing predictions. To make this difference concrete, we present a selection of the messages predicted as Negative by both systems in Table \ref{tab:predictions}. The first three are only predicted by the best ML system as Negative, and not by Pattern, while the fourth until the sixth examples are only seen as Negative by Pattern. Where the former give arguments (`can not be compared...’, `kids are dying from it’) or take stance (`I’m opposed to...’), the latter examples display more intensified words and exclamations (`that’s the message!!’, `Arrogant’, `horrific’) and aggression towards a person or organization. The last three tweets are seen by both systems as Negative. They are characterized by intensified words that linked strongly to a negative stance towards vaccination (`dangerous’, `suffering’, `get lost with your compulsory vaccination’).

Table \ref{tab:predictions} also features tweets that were predicted as Negative by neither the best ML-system nor Pattern, representing the most difficult instances of the task. The first two tweets include markers that explicitly point to a negative stance, such as `not been proven' and `vaccinating is nonsense'. 
The third tweet manifests a negative stance by means of the sarcastic phrase `way to go' (English translation). The use of sarcasm, where typically positive words are used to convey a negative valence, complicates this task of stance prediction. The last tweet advocates an alternative to vaccination, which implicitly can be explained as a negative stance towards vaccination. Such implicitly packaged viewpoints also hamper the prediction of negative stance. Both sarcasm and implicit stance could be addressed by specific modules.    

\subsection*{Improving recall}

For monitoring the number of Twitter messages over time that are negative towards vaccination, it is arguably more important to detect them at a high recall than at a high precision. False positives (messages incorrectly flagged as Negative) could be filtered manually by a human end user, while False Negatives (messages with a negative stance that are not detected) will be missed. We set out to improve recall, making use of classifier confidence scores and the complementary classifications of Pattern and the best ML system. 
%
%

A first recall-improving approach is to reset the prediction threshold for the Negative category. For any given instance, the SVM classifier estimates the probability of all categories it was trained on. It will predict the Negative category for an instance if its probability exceeds the probabilities of the other categories. This prediction can be altered by changing the threshold; setting the threshold higher will generally mean that fewer instances will be predicted as a Negative category (corresponding to a higher precision), whereas setting it lower will mean more instances will be predicted as such (corresponding to a higher recall).  Thus,  the balance between precision and recall can be set as desired, to favor one or another.  However, in many cases, changing the threshold will not lead to a (strong) increase in overall performance.

\begin{table}[t!]
\caption{Examples of tweets that were classified by the best ML system and/or pattern as `Negative’ (for privacy reasons, user mentions are replaced with `@USER').}
\begin{tabular}{L{8cm}|L{2cm}}
\hline
	Tweet (translated from Dutch) & Predicted as `Negative' by... \\ \hline
	 @USER aluminum which is a natural component in food cannot be compared to the stuff they put in that vaccine & ML only \\ 
	 @USER Kids are dying from it, what will you say to parents who are forced into inoculation despite their reluctance? & ML only \\
	 @USER And I'm opposed to having teenaged girls vaccinated against cervical cancer. @USER @USER @USER & ML only \\ \hline
	 @USER If your child is autistic after a vaccination, does the phrasing matter? No vaccinations, that's the message!! & Pattern only \\
	 @USER My experience with the RIVM is that I (mother) had proof that the inoculation was a trigger for epi. Arrogant and not empathic! @USER & Pattern only \\
	 @USER @USER I will never get inoculated again since this horrific experience \#scream \#connythemartyr & Pattern only \\ \hline
	 @USER True. But the inoculation is just like that. Dangerous junk & ML and Pattern \\
	Paternalistic bullshit. I had the measles, the mumps, Rubella and the fifth disease and I'm still here. Get lost with your COMPULSARY inoculation. & ML and Pattern \\
	The suffering called \#vaccination... \#nightparents 2.0 today... \#poor \#baby & ML and Pattern \\ \hline
    @USER Prevalence HPV is very low; effect has not been proven, extremely high frequency of medical issues after vaccination; simply criminal. & Neither ML nor Pattern \\
    Vaccinating is nonsense because polio is non-existent. & Neither ML nor Pattern \\
%
%
    Narcolepsy due to the vaccine against the swine flu. Way to go... \#eenvandaag & Neither ML nor Pattern \\
    Preventive colonoscopy saves many more lives than inoculating against virus cervical cancer 13-year olds. & Neither ML nor Pattern \\ \hline
\end{tabular}
\label{tab:predictions}
\end{table}

\begin{figure}[t!]
\includegraphics{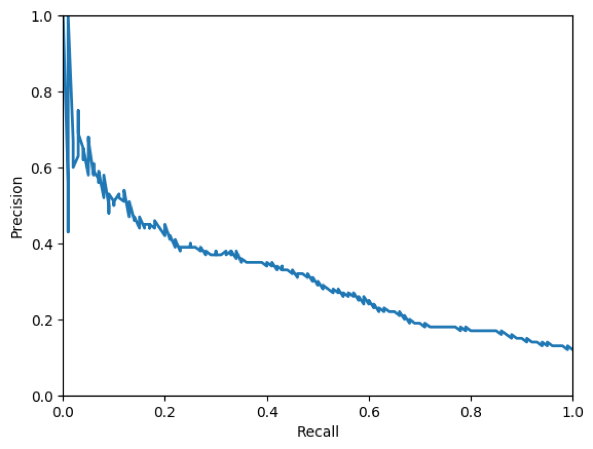}
\caption{\csentence{}Balance between precision and recall of predicting tweets with a negative stance when applying the best ML system, alternating the prediction threshold for this category.}
\label{fig:trade-off}
\end{figure}

Figure \ref{fig:trade-off} presents the balance between recall and precision as a result of predicting the Negative category with the best ML system, when the threshold for this category is altered from lowest to highest. Compared to the standard recall of $0.43$ at a precision of $0.29$, increasing the recall to $0.60$ would lead to a drop of precision to $0.21$. The F1-score would then decrease to $0.31$.   
%
%

A second means by which recall might be improved is to employ ensemble classification. The comparison in the previous section between the best ML method and rule-based sentiment analysis revealed that both systems have a rather disjoint perspective on negative stance: many more tweets are labeled as `Negative' by only one of the two systems than by both. We therefore built an ensemble system that follows both systems in their perspective on tweets with a negative stance: for each tweet, if either of the systems predicts the Negative category, the ensemble system makes this prediction.  
%
%

The performance of the ensemble system is presented in Table \ref{tab:ensemble}. Of the 343 tweets in the test set that are labeled as Negative, 210 are retrieved by the ensemble system. The result is a recall of $0.61$. The system does overshoot in its categorization of tweets as Negative: this category is predicted for 1,168 tweets (about 40\% of total test set of 2,886 tweets). The result is a precision of $0.18$. In comparison to lowering the prediction threshold of the ML system, the ensemble system thus yields a slightly worse trade-off between precision and recall.  

\begin{table}[h!]
\caption{Performance of the ensemble system on correctly predicting tweets labeled as `Negative' (AUC = Area under the curve, Total = all tweets in the test set that are labeled as `Negative', Predicted = the number of tweets that were classified as `Negative' by the system, Correct = the number of tweets that were correctly classified as `Negative'.}
      \begin{tabular}{rrrrrrr}
        \hline
        Precision & Recall & F1-score & AUC & Total	& Predicted & Correct \\ \hline
		0.18 & 0.61 & 0.28 & 0.62 & 343 & 1168 & 210 \\ \hline
      \end{tabular}
      \label{tab:ensemble}
\end{table}

\section*{Discussion}

With an F1-score of $0.36$, our system lags behind the $0.75$ F1-score reported by Du et al.\cite{Du+17}. Several factors might have influenced this difference. A first factor is the low proportion of tweets with the label `Negative' in our dataset. In the strict labeling condition, only 343 cases are labeled as negative by two annotators, against 2,543 labeled as positive -- the negative cases only comprise 13\% of all instances. In the study of Du et al., the anti-vaccination category comprises 24\% of all instances (1,445 tweets). More (reliable) examples might have helped in our study to train a better model of negative tweets. Secondly, Du et al. \cite{Du+17} focused on the English language domain, while we worked with Dutch Twitter messages. The Dutch Twitter realm harbors less data to study than the English one, and might bring forward different discussions when it comes to the topic of vaccination. It could be that the senders' stance towards vaccination is more difficult to pinpoint within these discussions. In line with this language difference, a third prominent factor that might have led to a higher performance in the study of Du et al.\cite{Du+17} is that they focus on a particular case of vaccination (e.g.: HPV vaccination) and split the anti-vaccination category into several more specific categories that describe the motivation of this stance. The diverse motivations for being against vaccination are indeed reflected in several other studies that focus on identifying discussion communities and viewpoints \cite{Bello-orgaz+17,Surian+16,Tangherlini+16}. While splitting the data into more specific categories will lead to less examples per category, it could boost performance on predicting certain categories due to a larger homogeneity. Indeed, the most dominant negative category in the study by Du et al.\cite{Du+17}, dubbed `NegSafety' and occurring in 912 tweets (63\% of all negative tweets), yielded the highest F1-score of $0.75$. While two less frequent categories were predicted at an F1-score of $0.0$, this outcome shows the benefit of breaking down the motivations behind a negative stance towards vaccination. 
%
%
	
A major limitation of our study is that the agreement rates for all categorizations are low. This is also the case in other studies, like \cite{Huang+17}, who report an agreement of $K = 0.40$ on polarity categorization. Foremost, this reflects the difficulty of the task. The way in which the stance towards vaccination is manifested in a tweet depends on the author, his or her specific viewpoint, the moment in time at which a tweet was posted, and the possible conversation thread that precedes it. Making a judgment solely based on the text could be difficult without this context. Agreement could possibly be improved by presenting the annotator with the preceding conversation as context to the text. Furthermore, tweets could be coded by more than two annotators. This would give insight into the subtleties of the data, with a graded scale of tweets that clearly manifest a negative stance towards vaccination to tweets that merely hint at such a stance. Such a procedure could likewise help to generate more reliable examples to train a machine learning classifier.

The low agreement rates also indicate that measuring stance towards vaccination in tweets is a too difficult task to assign only to a machine. We believe that the human-in-the-loop could be an important asset in any monitoring dashboard that focuses on stance in particular discussions. The system will have an important role in filtering the bigger stream of messages, leaving the human ideally with a controllable set of messages to sift through to end up with reliable statistics on the stance that is seen in the discussion at any point in time. In the analysis section, we explored two approaches to increase recall of messages with a negative stance, which would be most useful in this scenario. Lowering the prediction threshold showed to be most effective to this end.  

Our primary aim in future work is to improve performance. We did not experiment with different types of features in our current study. Word embeddings might help to include more semantics in our classifier’s model. In addition, domain knowledge could be added by including word lists, and different components might be combined to address different features of the data (e.g.: sarcasm and implicit stance). We also aim to divide the negative category into the specific motivations behind a negative stance towards vaccination, like in the study of Du et al.\cite{Du+17}, so as to obtain more homogeneous categories. Parallel to this new categorization of the data, adding more labeled data appears to be the most effective way to improve our model. The learning curve that we present in Figure \ref{fig:learning_curve} shows that there is no performance plateau reached with the current size of the data. An active learning setting \cite{Tong+01}, starting with the current system, could be applied to select additional tweets to annotate. Such a setting could be incorporated in the practical scenario where a human-in-the-loop judges the messages that were flagged as displaying a negative stance by the system. The messages that are judged as correctly and incorrectly predicted could be added as additional reliable training data to improve upon the model. We have installed a dashboard that is catered for such a procedure,\footnote{\url{http://prikbord.science.ru.nl/}} starting with the machine learning system that yielded the best performance in our current study.  
%
%


\section*{Conclusions}

We set out to train a classifier to distinguish Twitter messages that display a negative stance towards vaccination from other messages that discuss the topic of vaccination. Based on a set of 8,259 tweets that mention a vaccination-related keyword, annotated for their relevance, stance and sentiment, we tested a multitude of machine learning classifiers, alternating the algorithm, the reliability of training data and the labels to train on. The best performance, with a precision of $0.29$, a recall of $0.43$, an F1-score of $0.36$ and an AUC of $0.66$, was yielded by training an SVM classifier on strictly and laxly labeled data to distinguish irrelevant tweets and polarity categories. The baselines, with an optimal F1-score of $0.25$ (rule-based sentiment analysis), were considerably outperformed. The latter shows the benefit of machine-learned classifiers on domain-specific sentiment: despite being trained on a reasonably small amount of data, the machine-learning approach outperforms general-purpose sentiment analysis tools.

\section*{Availability and requirements}

{\bf Project name:} Prikbord \\
{\bf Project home page:} http://prikbord.science.ru.nl/  \\     
{\bf Operating system:} Linux \\
{\bf Programming language:} Python, javascript \\
{\bf Other requirements:} Django 1.5.11 or higher, MongoDB 2.6.10, pymongo 2.7.2 or higher, requests 2.13.0 or higher \\
{\bf License:} GNU GPL \\
{\bf Any restrictions to use by non-academics:} licence needed \\

\section*{Abbreviations}

{\bf EMM:} Europe Media Monitor 
{\bf MMR:} Mumps, Measles, Rubella 
{\bf LDA:} Latent Dirichlet Allocation 
{\bf ML:} Machine learning 
{\bf SVM:} Support Vector Machines 
{\bf AUC:} Area under the ROC Curve 
{\bf Clf:} Classifier 
{\bf NB:} Naive Bayes 
{\bf Pr:} Precision 
{\bf Re:} Recall 


\begin{backmatter}

\section*{Declarations}

\subsection*{Ethics approval and consent to participate}

Not applicable.

\subsection*{Consent for publication}

Not applicable.

\subsection*{Availability of data and materials}
http://cls.ru.nl/\~{}fkunneman/data\_stance\_vaccination.zip

\subsection*{Competing interests}
The authors declare that they have no competing interests.

\subsection*{Funding}
This study has been funded by the Rijksinstituut voor Volksgezondheid en Milieu. 

\subsection*{Author's contributions}
FK has set up the annotations procedure, performed the Machine Learning experiments and analysis, annotated tweets in the analysis and did a major part of the writing. ML has done part of the writing in the Introduction and Conclusion sections. AW has advised on the experimentation and analysis. AB has advised on the experimentation and has edited the complete text. LM has set up the annotations procedure, annotated tweets in the analysis and has done a major part of the writing. All authors read and approved the final manuscript.

\subsection*{Acknowledgements}
We thank Erik Tjong Kim Sang for the development and support of the \url{http://twiqs.nl} service. We also thank the ones who have contributed with annotations.

\bibliographystyle{bmc-mathphys} 
\bibliography{bmc_article}      


\section*{Figure legends}
  \subsection*{Figure \ref{fig:learning_curve}}
  
  \subsection*{Figure \ref{fig:trade-off}}
  
\end{backmatter}
\end{document}